\documentclass{article}
\usepackage{spconf,amsmath,graphicx, multirow}

\usepackage{enumitem}
\setlist{nosep, leftmargin=14pt}

\usepackage{mwe} 

\title{A Unified Framework for Semi-Supervised Image Segmentation and Registration}
\name{%
  Ruizhe Li$^{1}$ \quad
  Grazziela Figueredo$^{2}$ \quad
  Dorothee Auer$^{2}$ \quad
  Rob Dineen$^{2}$ \quad
  Paul Morgan$^{2}$ \quad
  Xin Chen$^{1}$%
}

\address{$^{1}$School of Computer Science, University of Nottingham, UK\\
        $^{2}$School of Medicine, University of Nottingham, UK}

\begin{document}
%
\maketitle
\begin{abstract}

Semi-supervised learning, which leverages both annotated and unannotated data, is an efficient approach for medical image segmentation, where obtaining annotations for the whole dataset is time-consuming and costly. Traditional semi-supervised methods primarily focus on extracting features and learning data distributions from unannotated data to enhance model training. In this paper, we introduce a novel approach incorporating an image registration model to generate pseudo-labels for the unannotated data, producing more geometrically correct pseudo-labels to improve the model training. Our method was evaluated on a 2D brain data set, showing excellent performance even using only 1\% of the annotated data. The results show that our approach outperforms conventional semi-supervised segmentation methods (e.g. teacher-student model), particularly in a low percentage of annotation scenario. GitHub: https://github.com/ruizhe-l/UniSegReg.

\end{abstract}

\vspace{-0.1cm}
\section{Introduction}
\vspace{-0.1cm}
\label{sec:intro}
Medical image segmentation is essential for various healthcare applications, including disease diagnosis, treatment planning, etc. Although supervised learning has shown great potential to automate segmentation tasks with high accuracy, it requires extensive annotated data for training, which is both time-consuming and costly to obtain. To overcome this limitation, semi-supervised learning \cite{yang2022survey} has emerged as a powerful approach, leveraging both annotated and unannotated data to train segmentation models without the need for extensive manual annotations. 

A common technique in semi-supervised learning is consistency regularization, which assumes that small perturbations to the data should not change its class label. One notable method is the Mean Teacher \cite{tarvainen2017mean}, which uses a teacher-student model in which the teacher model guides the student model using consistency loss to align their predictions. The Dual Students model \cite{ke2019dual} improves upon this by using two student models with different initial weights, enhancing stability and reducing bias. Such models have achieved better results in the field of data classification. In addition, Cui et al. \cite{cui2019semi} extended the model to the image segmentation area by adding a decoder to the network.

Pseudo-labeling is another approach in semi-supervised learning, where a model trained on annotated data generates labels for unannotated data to improve model robustness. However, generating accurate pseudo-labels is still challenging, as errors can persist. In image segmentation, evaluating confidence and selecting reliable pseudo-masks are particularly difficult. To address this challenges, Sun et al. \cite{sun2020teacher}, used teacher models to generate pseudo-masks, while Filipiak et al. \cite{filipiak2022polite} filtered out noisy labels using bounding boxes and mask scoring. Feng et al. \cite{feng2020semi} proposed dynamic mutual training, which leverages inter-model differences to correct errors, achieving good results. Li et al. \cite{li2020generic} proposed a semi-supervised approach that refines pseudo-labels iteratively using an ensemble of several submodels. Over time, the number of ensemble models is reduced, with each iteration improving the quality of the pseudo-labels to enhance segmentation performance. Bai et al. \cite{bai2017semi} introduced a self-learning approach that refines pseudo-masks using post-processing conditional random field and iterates the model until convergence. The above methods refine the pseudo-labels through different strategies, which help to improve the accuracy of segmentation even in the case of limited annotated data.

Alternatively, Spatial Transformer Network (STN)-based image registration models \cite{jaderberg2015spatial} have demonstrated notable success in medical image registration. These models can adaptively generate segmentation masks for the target image using the mask of the source image during the registration process. This capability helps retain more geometrically relevant information, addressing the issue of entirely erroneous pseudo-labels and offering a potential solution to generate high-quality pseudo-labels.

\begin{figure*}[!htb]
    \centering
    \includegraphics[width=17cm]{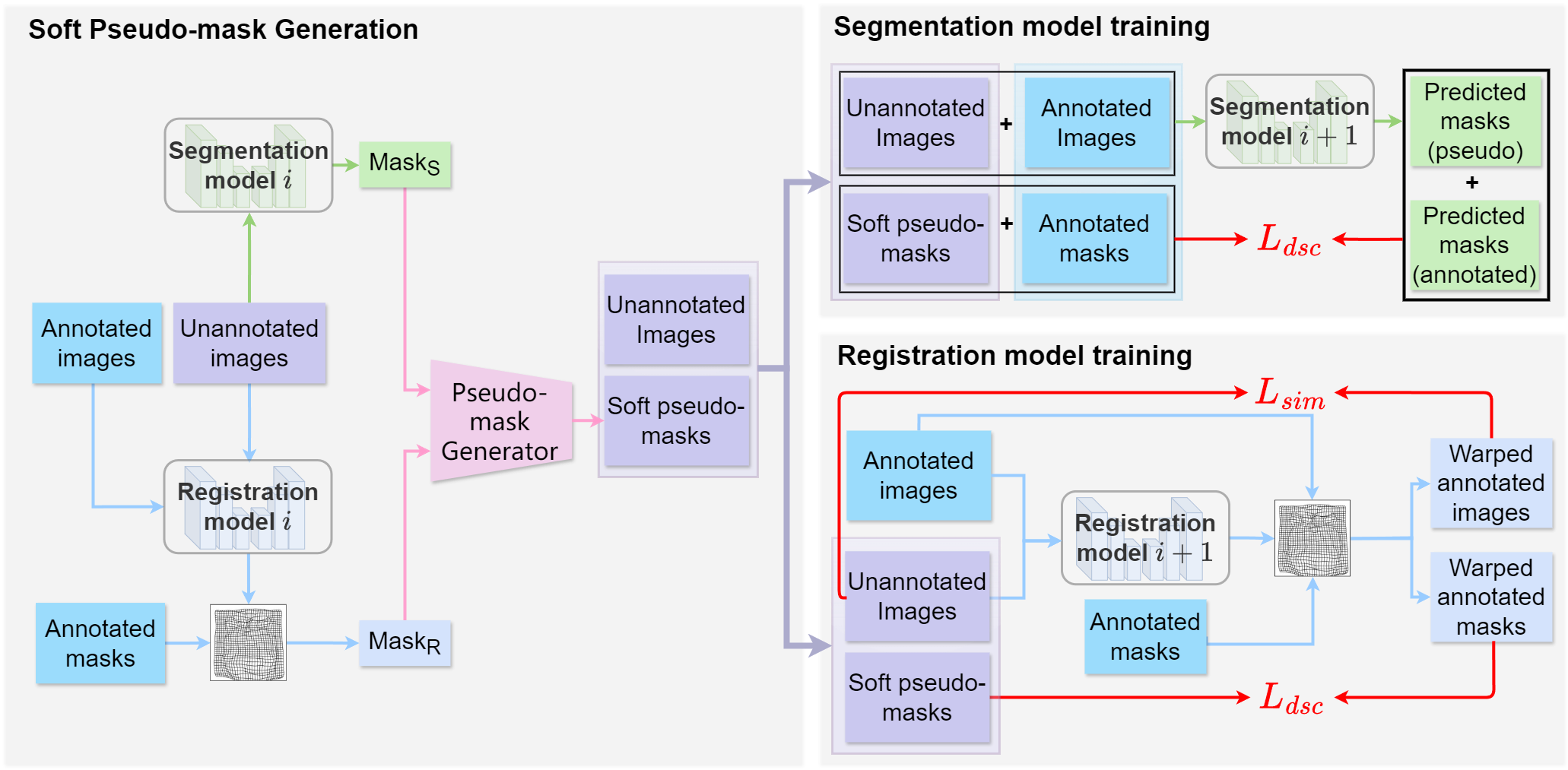}
    \vspace{-0.3cm}
    \caption{The proposed joint training framework for one iteration. It consists of soft pseudo-mask generation, segmentation model training, and registration model training. Pseudo-mask generation merges masks from both models for unannotated images to create soft pseudo-masks. These, with annotated images, form a new training set to enhance both models.}
    \label{fig:framework}
    \vspace{-0.4cm}
\end{figure*}

Inspired by image registration models and the pseudo-labeling methods, in this paper, we introduce an integrated image segmentation and registration framework with a novel component called ``soft pseudo-mask generation". The contributions of the proposed method are summarized as follows. (1) We propose an iteratively trained framework that integrates image segmentation and registration methods, showcasing significant improvements in both image registration and segmentation in a medical imaging datasets with a small number of annotated images. (2) A soft pseudo-mask generation process is introduced, which enhances the quality of pseudo-masks generated from the segmentation and registration models, reducing the impact of incorporating locally incorrect pseudo-mask information. The generated soft pseudo-mask can also act as a confidence map in the final segmentation prediction, benefiting downstream tasks.
\vspace{-0.3cm}
\section{Methodology}
\vspace{-0.2cm}
\label{sec:methods}

\subsection{Overall Framework Architecture}
\label{ssec:methods-model}

\noindent Fig. \ref{fig:framework} illustrates the proposed framework to achieve joint image segmentation and registration. The segmentation and registration models are trained iteratively. Initially, a U-Net \cite{ronneberger2015u} based segmentation model with added residual blocks is trained on annotated images, while an unsupervised registration model based on our previously developed MrRegNet \cite{li2024mrregnet} is trained on both annotated and unannotated images without using segmentation masks. Both segmentation and registration models then generate pseudo-masks for unannotated images, which are used for subsequent training iterations. 

This process allows the segmentation and registration models to progressively enhance each other. Specifically,  additional pairs of unannotated images and pseudo-masks are used to retrain the segmentation model, while the pseudo-masks are also used to guide the image registration model to focus more on the segmented region of interest (ROI). 

\vspace{-0.1cm}
\subsection{Soft Pseudo-masks Generation}
\vspace{-0.1cm}

Besides the segmentation and registration models, soft pseudo-mask generation is another key component of the proposed method. The segmentation model generates $N$ probabilistic pseudo-masks using test-time data augmentation, including rotation, flipping, and contrast adjustment. The registration model generates a displacement field that maps an annotated source image to an unannotated target image, which is then used to warp the source mask to the target coordinate, producing a pseudo-mask for the unannotated image. For each unannotated image, $N$ source images are used to generate $N$ pseudo-masks through the registration model. 

The final soft pseudo-mask for each unannotated image is obtained by averaging all $2N$ generated pseudo-masks from both models. The term ``soft" means that, instead of a binary mask, pixel values in the soft mask range between 0 and 1, representing confidence scores. These unannotated images, along with their soft pseudo-masks, are then used to train both models in the next iteration. This iterative process gradually enhances the quality of the soft pseudo-masks, reducing the risk of the models learning incorrect information from inaccurate masks. Fig. \ref{fig:vis-iter} in section \ref{ssec:experiments-results} shows this process. 

\vspace{-0.1cm}
\subsection{Model Training} 

\textbf{Segmentation model:} The segmentation model (i.e. residual U-Net \cite{ronneberger2015u}) is initially trained using annotated images, and in the following iterations the unannotated images with their generated pseudo-masks are used for training. Optimized for smoother convergence, the soft Dice loss is employed to improve model performance by leveraging soft pseudo-masks. Additionally, following the approach outlined in the V-Net \cite{milletari2016v}, squaring is applied to the denominator of the loss function (${y'}_{p}^2$ and $y_{p}^2$) to create a smoother landscape for faster convergence. The equation for the soft Dice loss function is expressed as follows: 
\begin{equation}
    \label{eq:softdice}
    L_{dsc}(y', y) = 1-\frac{2\sum_{p}^{\Omega}({y'}_p y_p)}{\sum_{p}^{\Omega}({y'}_{p}^2)+\sum_{p}^{\Omega}(y_{p}^2)}
\end{equation}
where $p$ indicates the index of pixels in the whole image $\Omega$. $y$ and $y'$ represent the annotated mask (or soft pseudo mask) and the predicted segmentation map, respectively.

\textbf{Registration model:} The loss function in the MrRegNet registration model \cite{li2024mrregnet}) uses global normalized cross-correlation (GNCC) to measure the similarity between the warped source image and the target image, as described in Eq. (\ref{eq:sim}):
\begin{equation}
    \label{eq:sim}
    L_{sim}(x, y)=-\frac{1}{N}\sum_{p \in \Omega}\frac{(x_p-\overline{x})(y_p-\overline{y})}{\sigma_{x}\sigma_{y}} 
\end{equation}
where $x$ and $y$ are the warped source image and the target image respectively. $\overline{x}$ and $\overline{y}$ are the mean intensity values of $x$ and $y$ while $\sigma_{x}$ and $\sigma_{y}$ denoting their variances.

In addition, annotated masks and pseudo masks are utilized to guide the model in focusing on the masked ROI using the soft Dice loss in Eq.(\ref{eq:softdice}), along with an $L2$ smoothness regularization term to regularize the displacement field. In summary, the image registration model is trained by optimizing the objective function in Eq. (\ref{eq:opt-reg}). 
\begin{multline}
    \label{eq:opt-reg}
    \underset{D}{arg \; min} \, \frac{1}{K}\sum_{i=1}^{K}(L_{sim}(f_{D_i}(x_S), x_T) +  L_{dsc}(f_{D_i}(y_S), y_T) \\ + \lambda \left \| \bigtriangledown D_i \right \|)
\end{multline}
where $D$ represents the displacement field, and $D_i$ denotes the final displacement field at each resolution out of $K$ resolutions, as in MrRegNet \cite{li2024mrregnet}, the displacement field is estimated in a multi-resolution manner. $x_S$ and $x_T$ indicate the source and target images, while $y_S$ and $y_T$ represent the source and target masks, respectively. Finally, $f_{D}(x_S)$ and $f_{D}(y_S)$ are the warped source image and the warped source mask respectively.

\section{Experiments and results}
\label{sec:experiments}

\subsection{Dataset}
\label{sec:dataset}

The proposed semi-supervised segmentation methods were evaluated on a local 2D brain MRI dataset comprising 820 T1-weighted slices, each with manually annotated mid-brain regions. This dataset included images from various subjects and MRI scanners, resulting in significant intensity and geometric variations that posed challenges for segmentation. The data was split into training (80\%) and testing (20\%) sets, with 620 images for training and 200 for testing; an additional 20 images from the training set were allocated for validation. All images were resized to $256\times256$ pixels and normalized between 0 and 1 using min-max normalization, ensuring consistent pre-processing across experiments.

\subsection{Experimental Design and Parameter Settings}
\label{ssec:experiments-design}

Experiments were conducted with varying amounts of annotated data—1\% (5 images), 10\% (60 images), and 50\% (300 images)—to simulate data-scarce scenarios. The fully supervised (FS) model, trained on different amounts of annotated images, served as a baseline method. Two semi-supervised methods, the widely used Mean Teacher (MT) model \cite{sun2020teacher} and our proposed ``Joint'' model, were compared in this paper, both trained on a mix of annotated and unannotated data (e.g., MT-1\% and Joint-1\% used 1\% annotated images and 99\% unannotated data). The combined soft mask results (i.e. ``Combined") produced by our method, which integrate both registration and segmentation outputs, are also presented.

All models used consistent parameter settings. The segmentation network began with a learning rate of 0.0001 for 500 epochs, reduced to 0.00001 for 100 epochs in later iterations. The registration network’s initial learning rate of 0.001 (200 epochs) was later adjusted to 0.0001 (50 epochs). The registration network contained 5 levels of image resolution with the smoothness term weight ($\lambda_i$) progressively decreasing by half at each level, from $\lambda_1 = 128$ to $\lambda_5 = 8$.

Each model generated 5 pseudo-masks ($N=5$) per iteration for soft pseudo-mask generation. The same segmentation architecture was applied to the MT method, where student model weights were updated by the teacher model after 100 epochs using the exponential moving average (EMA) loss, over a total of 500 epochs.

\subsection{Results}
\label{ssec:experiments-results}

The segmentation results from different methods, evaluated by Dice Coefficient (DSC) and Hausdorff Distance (HD, in pixel units), are summarized in Table \ref{table:result-seg}. Higher DSC values and lower HD values indicate better segmentation accuracy and boundary correctness. The percentage values represent the amounts of annotated data used for model training.

\begin{table*}[htb!]
\centering
\caption{Segmentation results using different rates of annotated images. The mean $\pm$ standard deviation of DSC and HD (in pixels) for FS, MT, the proposed Joint model and the Combined soft mask are reported. Results between MT and Combined soft mask were statistically analyzed using the Wilcoxon Signed Rank test, with * indicating \( p < 0.05 \).}
\resizebox{\textwidth}{!}{%
\begin{tabular}{|l|cccc|cccc|}
\hline
\multicolumn{1}{|c|}{\multirow{2}{*}{\textbf{Method}}} & \multicolumn{4}{c|}{\textbf{Dice Coefficient (DSC)}} & \multicolumn{4}{c|}{\textbf{Hausdorff Distance (HD)}} \\
\cline{2-9}
 & \textbf{1\%} & \textbf{10\%} & \textbf{50\%} & \textbf{100\%} & \textbf{1\%} & \textbf{10\%} & \textbf{50\%} & \textbf{100\%} \\
\hline
FS & 0.58±0.27 & 0.83±0.15 & 0.92±0.05 & \textbf{0.93±0.02} & 32.76±45.33 & 14.44±22.39 & 10.06±21.40 & \textbf{8.43±19.70} \\
MT & 0.47±0.31 & 0.85±0.09 & 0.90±0.06 & n/a & 80.45±61.74 & 14.16±24.96 & 11.21±22.03 & n/a \\
Joint & 0.85±0.05 & 0.86±0.06 & 0.92±0.05 & n/a & 12.87±17.97 & 11.90±18.47 & 9.84±21.64 & n/a \\
Combined & \textbf{0.84±0.05}* & \textbf{0.87±0.04}* & \textbf{0.92±0.03}* & n/a & \textbf{11.60±18.57}* & \textbf{11.26±18.77}* & \textbf{8.99±19.56}* & n/a \\
\hline
\end{tabular}}
\label{table:result-seg}
\vspace{-0.3cm}
\end{table*}

\begin{figure}[!htb]
    \centering
    \includegraphics[width=\columnwidth]{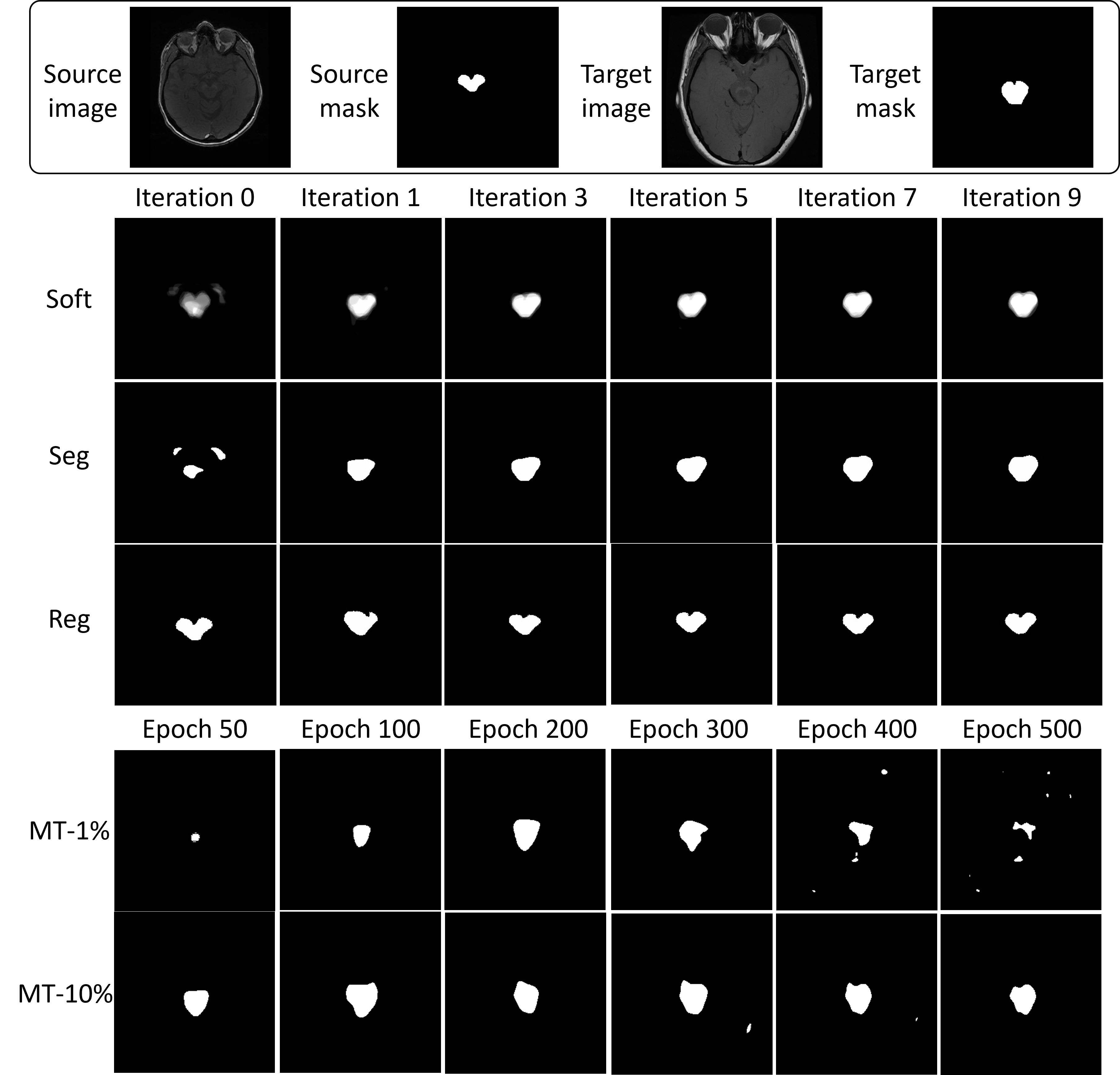}
    \caption{Visualization results of an example that participated in training as an unannotated image in Joint-1\% (row 1-3), MT-1\% (row 4) and MT-10\% (row 5).}
    \label{fig:vis-iter}
    \vspace{-0.1cm}
\end{figure}

The quantitative results in Table \ref{table:result-seg} show that at all rates of annotated data for training, both the joint model and the combined soft mask produce superior DSC values compared to the MT method. In particular, with only 1\% annotated data, the combined soft mask achieves a significantly higher DSC (0.85) and a lower HD (11.6) compared to the baseline model (DSC 0.58) and the MT model (DSC 0.47). The Joint model leverages pseudo-labels more effectively than the MT model when the amount of annotated data is extremely small (e.g. 1\%). The MT 1\% model's DSC is even lower than the baseline (0.47 v.s. 0.58). It is mainly due to the low-quality pseudo-labels generated during training, suggested by the qualitative results shown in Fig. \ref{fig:vis-iter}. As the amount of annotated data increases from 10\% to 50\%, our method keeps outperforming the MT method. 

In addition to the quantitative results, a qualitative analysis was conducted to observe the improvement of soft pseudo masks during training. Fig. \ref{fig:vis-iter} shows the pseudo masks of the target image generated by the segmentation and registration blocks of the Joint-1\% model in different iterations. For comparison, pseudo masks from the MT-1\% and MT-10\% models at various epochs are also shown.

At iteration 0 using our method, the pre-trained segmentation model shows large errors, and the pre-trained registration model (without mask guidance) achieves only a coarse alignment. Initially, the soft pseudo-mask shows a minimal overlap between the $2N$ pseudo-masks, but as training progresses, the geometric information contributed from the registration model refines the alignment and segmentation results. This integration allows the soft pseudo-mask to gradually evolve with higher and higher confidence. Both models iteratively enhance each other, with the Joint model converging rapidly by iteration 7. The final soft mask serves as a confidence map, incorporating the spatial consistency from the registration model.

In contrast, Fig. \ref{fig:vis-iter} row 4 shows that MT-1\% produced unreliable pseudo-masks due to limited annotated data, leading to degraded performance over epochs. Although Fig. \ref{fig:vis-iter} row 5 demonstrates improved stability by using MT-10\%, it remains less robust compared to our method. Additionally, the MT model lacks the ability to preserve the geometric information in their generated pseudo-masks.

In summary, the quantitative and qualitative results demonstrate that our joint model significantly improves segmentation accuracy through geometric preservation provided by the registration model in low-annotation-rate scenarios. The combined soft mask, a post-processed output of the joint model, further improves performance. This highlights the robustness and strength of the proposed methods in generating high-quality pseudo-masks, compared to the MT method.

\vspace{-0.1cm}
\section{Discussion and Conclusions}
\label{sec:conclusions}

This study presents a semi-supervised-learning framework for joint training of image segmentation and registration models using an iterative process. Initially, the segmentation model is trained on a small set of annotated data, while the registration model is trained in an unsupervised manner on both annotated and unannotated data. The framework leverages a soft pseudo-mask generation module to produce high-quality pseudo-masks, which help to iteratively improve both the segmentation and registration models. This approach has been shown to produce accurate segmentation and registration results, even with 1\% of annotated data while preserving anatomical structures.

The generated soft segmentation masks can also serve as confidence maps, useful for quality control and uncertainty estimation. Further improvements could be achieved by increasing the number of augmented images for segmentation and template source images for registration.

\section{Compliance with Ethical Standards}
This research study was conducted retrospectively using a local brain MRI dataset of human subjects. Ethical approval for the use of the brain MRI dataset was obtained. The subjects used in this study had consented to be included in this research. All data was anonymized, and the participants’ information cannot be identified from the imaging data.

\section{Acknowledgments}
\label{sec:acknowledgments}

This work was supported by the NIHR Nottingham Biomedical Research Centre (BRC) at the University of Nottingham. The authors sincerely appreciate the funding and resources provided, which have been instrumental in conducting this research.

\bibliographystyle{IEEEbib}
\bibliography{strings}

\end{document}